# From Sliding Windows to Self-evolving Lifelong Memory: A Dual-perspective Survey of Memory Paradigms for Long-Horizon Robotics VLA Models

Yong Wang [0009-0003-2943-4670], Boyu Shen [0009-0002-4801-0892], Xinwei Zhang [0009-0004-3582-3901], Yizheng Li [0009-0006-7475-8457] and Keyuan Luo *[0009-0006-5265-784X]

Department of Computer Science and Technology, Hainan Bielefeld University of Applied Sciences, Hainan 571700, China
keyuan.luo@hibiuh.edu.cn

**Abstract.** Long-horizon robotic tasks expose a fundamental limitation of current Vision-Language-Action (VLA) models because their restricted temporal context prevents effective accumulation and reuse of past experiences. As a result, memory has become a fundamental component for enabling embodied agents to perform long-horizon reasoning and adaptive decision-making. This survey presents a systematic taxonomy of VLA memory mechanisms based on memory scope and persistence, categorizing existing approaches into five representative paradigms and one emerging frontier. Based on this taxonomy, we compare representative VLA memory systems in terms of memory characteristics, architectural complexity, and open-source maturity, revealing a clear trade-off in which increasing architectural complexity is generally accompanied by decreasing open-source maturity. Finally, we discuss the key challenges in building full-lifecycle memory systems, including efficient memory management and lifelong knowledge evolution. Overall, this survey provides a unified perspective on memory architectures for VLA agents and offers insights into the development of lifelong embodied intelligence.

## 1 Introduction

Vision-Language-Action (VLA) models have demonstrated strong capabilities in mapping visual and linguistic inputs to robotic actions [1]. However, long-horizon manipulation remains challenging due to limited temporal context and insufficient mechanisms for storing and reusing task-relevant experiences [5, 8, 26].

Existing studies mainly investigate individual memory mechanisms, while a unified framework for characterizing memory evolution in VLA systems remains lacking [1, 8]. Unlike previous surveys that categorize VLA systems according to architectures or learning strategies [1], this survey adopts memory as the organizing principle and introduces a Scope–Persistence perspective to characterize the evolution from short-context interaction toward lifelong self-evolving memory. Existing memory mechanisms are further analyzed from the perspectives of memory evolution, architectural complexity, and practical accessibility. The analysis reveals a trade-off between increasing memory capability and decreasing accessibility. Persistent deployment-time memory and self-evolving memory remain open challenges due to the limited availability of full-stack implementations [10].

The main contributions of this survey are summarized as follows:

- We introduce a unified Scope–Persistence taxonomy to characterize the evolution of memory architectures in VLA systems, providing a memory-centric perspective beyond existing architecture- or learning-based categorizations.
- We systematically review existing VLA memory mechanisms and organize them into five established paradigms and one emerging frontier, covering their memory representations, retrieval strategies, and update mechanisms.
- We establish a unified comparison framework considering memory scope, memory form, persistence, speed, forgetting behavior, and open-source maturity, revealing the trade-off between memory capability and system accessibility.
- We identify self-evolving lifelong memory as an emerging direction and discuss key challenges toward developing persistent and continuously adaptive embodied agents.

## 2 Background and Taxonomy

### 2.1 VLA Models and Long-horizon Tasks

Vision-Language-Action (VLA) models integrate visual perception, language understanding, and robotic control to generate actions from multimodal inputs [1, 30]. Related multimodal learning strategies have also been explored in other embodied intelligent systems, where transformer-based architectures exploit cross-modal interactions between visual and contextual information for perception and decision-making [31–33]. However, existing VLA policies often rely on limited temporal context, restricting their ability to perform long-horizon tasks requiring progress tracking, failure recovery, and experience reuse [1, 2].

### 2.2 Dimensions of Comparison

We adopt six dimensions to compare memory paradigms, including Scope, Memory Form, Persistence, Speed, Forgetting, and Open-source Maturity (OSS). Scope describes the temporal and contextual span of accessible experience, while Memory Form characterizes its representation, including implicit parameter-based memory and explicit external memory [3,7]. Persistence indicates whether experience can be retained beyond the current episode [3]. Speed, Forgetting, and OSS evaluate deployment efficiency, knowledge retention, and implementation accessibility, respectively.

Beyond these dimensions, memory systems are analyzed through four lifecycle mechanisms, including representation, retrieval, updating, and consolidation [3, 5, 7]. To improve the consistency of qualitative comparison, each dimension is evaluated according to predefined criteria. Scope evaluates the range of available experience from recent observations to lifelong knowledge. Speed considers retrieval latency and computational overhead during deployment. Forgetting reflects the ability to preserve previously acquired knowledge during future tasks. OSS maturity considers the availability of source code, pretrained models, documentation, and reproducible implementations.

The qualitative ratings in Table 1 are determined based on relative comparisons among representative systems rather than absolute measurements. For example, speed reflects additional computational overhead introduced by memory retrieval and management, while OSS maturity considers the availability of public implementations, pretrained models, and reproducible experimental settings.

### 2.3 The Memory Paradigms

We propose a taxonomy of VLA memory mechanisms based on two complementary dimensions, namely memory scope and cross-episode persistence. Memory scope characterizes the temporal and contextual extent of information available to the agent, whereas cross-episode persistence reflects whether acquired knowledge persists beyond a single episode [3]. These two dimensions provide a model-independent perspective for analyzing the evolution of VLA memory mechanisms.

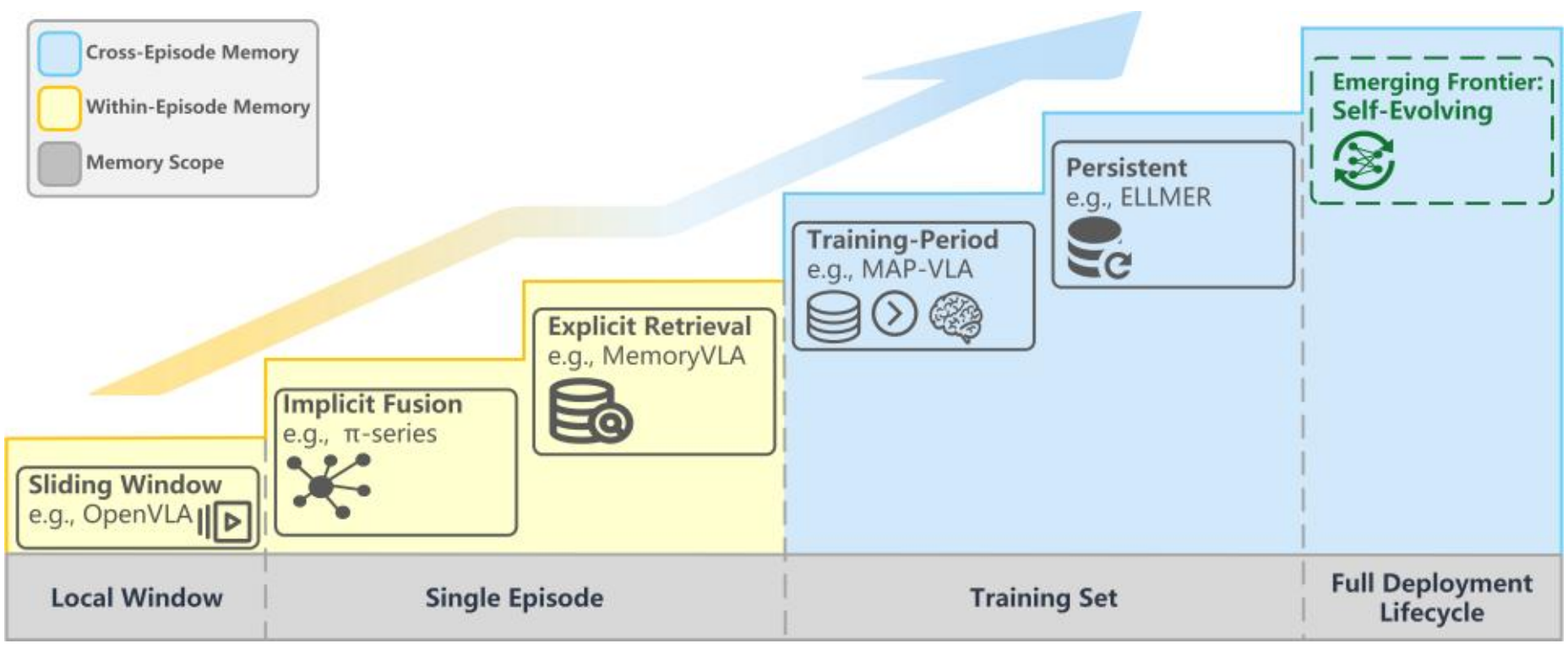

**Fig. 1.** Scope–Persistence taxonomy illustrating the evolution of representative VLA memory paradigms across different memory scopes and persistence levels.

Existing memory paradigms can be organized according to memory scope and persistence, as shown in Fig. 1. Rather than representing a strictly linear progression, different paradigms emerge to address distinct memory requirements. Based on their defining characteristics, existing approaches are categorized into six representative types: sliding window memory, implicit fusion memory, explicit retrieval memory, training-period memory, persistent memory, and self-evolving memory.

**Table 1.** Qualitative comparison of representative VLA memory paradigms.

| Paradigm | Scope | Memory Form | Persistence | Speed | Forgetting | OSS | Examples |
|---|---|---|---|---|---|---|---|
| Sliding Window | 1-6 frames | Implicit | None | Highest | Excellent | High | OpenVLA |
| Implicit Fusion | Episode | Implicit | None | High | Medium | High | OpenPI, $\pi_0$ |
| Explicit Retrieval | Episode | Explicit | None | Moderate | Medium | Medium | MemoryVLA |
| Training Period | Training | Weights | Train only | High | Poor | Medium | MAP-VLA |
| Persistent | Deployment | Explicit | Full | Moderate | Excellent | Low | ELLMER |
| Self-evolving | Lifecycle | Unified | Full | N/A | N/A | Concept | No mature implementation |

**Table 2.** Engineering Mechanisms Underlying Different Memory Paradigms.

| Paradigm | Memory Representation | Retrieval Strategy | Update and Consolidation |
|---|---|---|---|
| Sliding Window | Observation sequence | Context attention | None |
| Implicit Fusion | Latent state | Implicit aggregation | Learned update |
| Explicit Retrieval | Episodic memory bank | Similarity-based retrieval | Manual writing |
| Training Period | Model parameters | Parameter inference | Offline optimization |
| Persistent | Persistent memory store | Cross-episode retrieval | Memory management |
| Self-evolving | Prospective unified memory | Prospective autonomous retrieval | Prospective autonomous evolution |

The first three paradigms focus on within-episode reasoning and differ mainly in how historical information is retained, ranging from temporary context windows to latent representations and explicit memory retrieval. The remaining paradigms extend memory across episodes by enabling knowledge reuse over time. This evolution reflects a gradual transition from short-term context utilization to persistent and adaptive memory systems.

Representative systems are compared using the framework introduced in Section 2.2. Table 1 summarizes their memory characteristics and practical accessibility, while Table 2 further analyzes the underlying engineering mechanisms. These comparisons reveal a trade-off between memory capability and implementation complexity. More advanced memory mechanisms generally provide broader temporal coverage and stronger knowledge retention, but often introduce higher computational requirements and lower reproducibility.

## 3 Within-episode Memory

Within a single task execution, memory mechanisms enable the agent to make short-term decisions by leveraging recent observations, latent representations, or retrieved experiences [2]. The policy parameters are kept unchanged during deployment, and memory is instead maintained through temporary context, latent states, or external retrieval mechanisms.

### 3.1 Sliding Window

In sliding-window policies, the model only uses a limited number of recent observations with task instructions for action prediction rather than maintaining complete episode history. This approach follows sequence modeling paradigms such as Decision Transformer [29] and is adopted by VLA policies including OpenVLA [16], Octo [18], and RoboFlamingo [11], as well as visuomotor policies such as Diffusion Policy [24]. Since earlier observations are discarded after leaving the input window, these methods cannot reuse past experiences across episodes. This design reduces computational overhead and avoids additional memory management during deployment, but prevents the policy from revisiting potentially important historical information. Because model parameters remain fixed after training, new experiences cannot be integrated during execution. Therefore, sliding-window memory achieves efficient short-term reasoning at the cost of limited temporal coverage and poor long-horizon adaptability. Extending the context window can partially alleviate this limitation but introduces higher inference cost and memory consumption [3, 5, 8]. This creates a practical trade-off between short-term efficiency and long-horizon information retention, making sliding-window approaches less suitable for tasks requiring delayed reasoning or recovery from earlier failures.

### 3.2 Single-episode Implicit Fusion

Recent VLA models, including π0 [13] and OpenPI [14], typically do not maintain an independent persistent memory module during deployment. Instead, historical information is temporarily retained through latent representations or extended context windows, enabling the policy to incorporate previous observations when generating actions. Since inference does not update model parameters, such information cannot be preserved as reusable memory beyond the current interaction [3]. Although implicit fusion improves within-episode temporal reasoning, it suffers from limited interpretability because the model cannot explicitly identify or modify retained experiences. Moreover, extending context length increases computational costs, creating a transparency–efficiency trade-off that limits its applicability to long-horizon experience reuse [1, 3, 8].

### 3.3 Single-episode Explicit Retrieval

Explicit retrieval-based memory stores previous task experiences in external memory and retrieves relevant information during execution. MemoryVLA [10], MemER [9], KEMO [27], and Embodied-RAG [28] follow this strategy by separating memory storage from policy inference, enabling experience reuse without updating policy parameters. Compared with implicit fusion, retrieval-based memory provides more interpretable and accessible stored information [1]. However, growing memory banks increase retrieval latency and storage costs, while irrelevant or outdated experiences may introduce retrieval noise [3,8–10]. Moreover, retrieval alone cannot update the underlying policy or consolidate new knowledge, limiting continuous adaptation.

Therefore, retrieval-based approaches trade flexible experience access for increased memory management complexity.

## 4 Cross-episode Memory

### 4.1 Training-period Cross-episode Enhancement

Paradigm 4 integrates accumulated experience during training or fine-tuning, consolidating task-relevant knowledge into model parameters through optimization. Representative systems include MAP-VLA [19] and GR-2 [4], which leverage training-stage experience to improve generalization. After training, the acquired knowledge is retained in fixed parameters, distinguishing this paradigm from deployment-time persistent memory. However, new experiences cannot be incorporated without additional optimization, requiring retraining or fine-tuning that increases computational cost and may cause catastrophic forgetting of previously acquired skills.

### 4.2 Deployment-time Persistent Memory

Deployment-time persistent memory architectures maintain task-relevant information beyond individual episodes through external memory structures, separating memory from policy parameters and enabling retrieval during deployment. Recent embodied AI systems, including ELLMER [15], RoboMemory [17], KARMA [25], and ECHO [21], explore persistent memory through external storage, hierarchical organization, and consolidation mechanisms. However, current approaches mainly focus on memory storage and retrieval, while autonomous organization, knowledge consolidation, and adaptive updating remain open challenges. Persistent memory also introduces a stability–plasticity trade-off: frequent updates improve adaptation but risk overwriting valuable knowledge, whereas conservative updates limit flexibility. In addition, large-scale memory stores increase storage and retrieval costs, hindering long-term robot deployment.

### 4.3 Full-lifecycle Multimodal Self-evolving Memory

Paradigm 6 represents an emerging direction toward self-evolving lifelong memory for VLA agents. Unlike persistent memory mechanisms that mainly store and retrieve experiences, this paradigm aims to enable autonomous memory updating, knowledge consolidation, and capability refinement through continuous interaction [1]. However, fully autonomous self-evolving memory has not yet been demonstrated in current VLA systems [1]. Achieving this capability requires balancing adaptation and knowledge stability while addressing challenges in memory reliability, scalability, and autonomous memory management [3].

## 5 Benchmarks and Evaluation

Recent benchmarks evaluate different aspects of embodied intelligence, including manipulation generalization through SimplerEnv [20], continual skill learning through LIBERO and LIBERO-Long [8], long-horizon execution through CALVIN [2], compositional reasoning through LoHoRavens [22], and memory-intensive manipulation in non-Markovian environments through LIBERO-Mem [23]. However, existing evaluations still mainly focus on task-level metrics, such as success rate, execution duration, and continual learning performance, while providing limited assessment of memory quality and effectiveness [8]. In particular, current continual learning benchmarks mainly measure forgetting through performance degradation after sequential training, but rarely evaluate whether agents can retain semantic knowledge, manipulation skills, and previous experiences [3, 5, 6, 8, 12]. Future benchmarks should therefore incorporate quantitative metrics for memory retention, forgetting resistance, retrieval effectiveness, and memory efficiency, considering the relationship between stored information, retrieval cost, and task performance.

Quantitative evaluation of memory capabilities should consider both task performance and memory-specific metrics. Memory retention can be measured by evaluating whether previously acquired skills remain accessible after long intervals or intervening tasks. Retrieval effectiveness can be evaluated through retrieval accuracy or recall of relevant experiences, while memory efficiency can consider storage consumption and inference latency. Such metrics would enable more direct comparison between different memory architectures beyond task success rates.

## 6 Open Challenges and Future Directions

VLA memory research has gradually expanded from using short observation contexts to storing and reusing information across longer interactions. However, current memory designs still lack a complete mechanism for continuous adaptation during robot deployment. Key challenges remain in memory organization, experience incorporation, knowledge maintenance, and efficient retrieval. Most existing methods separate memory storage from usage, leaving memory selection, modification, and forgetting strategies underexplored. For long-term operation, adaptive memory management based on task relevance, usage frequency, and future utility is needed to maintain compact and effective memory structures [12]. Moreover, persistent memory updating remains limited, as many approaches rely on training-time knowledge or temporary task records. Modular memory structures that enable deployment-time updates without retraining remain an important direction [10].

Beyond experience storage, lifelong VLA learning also requires transforming accumulated information into reusable skills. Memory consolidation and automatic skill discovery may help extract transferable behaviors from previous interactions [8, 15]. Meanwhile, efficient retrieval becomes increasingly important as multimodal memory grows, where hierarchical retrieval mechanisms can reduce search costs and improve access to relevant experiences in long-horizon tasks [9].

## 7 Conclusion

Memory in VLA models has been reviewed in this survey from the perspective of scope and persistence, providing a taxonomy that connects different memory architectures within a common framework. Based on representative systems covering sliding-window policies, implicit fusion, retrieval-based memory, and persistent memory designs, we discuss how memory mechanisms have developed to support long-horizon embodied intelligence. These approaches have improved the reuse of previous experiences and temporal reasoning ability, but persistent memory during deployment remains limited in current open-source VLA systems. The gap between algorithmic progress and practical deployment indicates the need for memory mechanisms that can retain, retrieve, consolidate, and update knowledge over long-term operation while satisfying computational constraints. The discussion in this survey is also subject to several limitations. The taxonomy and qualitative comparison are derived from the representative systems considered in this work, and the lack of standardized benchmarks still makes it difficult to evaluate different memory mechanisms in a unified manner. Broader evaluations covering more emerging VLA systems, together with quantitative protocols for memory retention, retrieval, and lifelong adaptation, will be important for future studies.